# A BAYESIAN GENERATIVE ADVERSARIAL NETWORK (GAN) TO GENERATE SYNTHETIC TIME-SERIES DATA, APPLICATION IN COMBINED SEWER FLOW PREDICTION

## Amin E. Bakhshipour[1], Alireza koochali[1,2,3], Ulrich Dittmer[1], Ali Haghighi[4], S. Ahmad[1,3], A. Dengel[1,3]

[1] TU Kaiserslautern, Kaiserslautern, Germany
[2] Ingenieurgesellschaft Auto und Verkehr(IAV) GmbH, Berlin,Germany
[3] German Research Center for Artificial Intelligence (DFKI), Kaiserslautern, Germany
[4] Shahid Chamran University of Ahvaz, Ahvaz, Iran

[1]  0000-0002-6921-2381, amin.bakhshipour@bauing-uni.kl.de, [2]  0000-0001-7370-9369, alireza.koochali@iav.de, [3]  0000-0003-1723-3356, ulrich.dittmer@bauing.uni-kl.de, [4]  0000-0002-2765-6929, a.haghighi@scu.ac.ir, [5]  0000-0002-4239-6520, sheraz.ahmed@dfki.de, [6]  0000-0002-6100-8255, andreas.dengel@dfki.de

## Abstract

Despite various breakthroughs of machine learning and data analysis techniques for improving smart operation and management of urban water infrastructures, some key limitations obstruct this progress. Among these shortcomings, the absence of freely available data due to data privacy or high costs of data gathering and the nonexistence of adequate rare or extreme events in the available data plays a crucial role. Here, the Generative Adversarial Networks (GANs) can help overcome these challenges. In machine learning, generative models are a class of methods capable of learning data distribution to generate artificial data. In this study, we developed a GAN model to generate synthetic time series to balance our limited recorded time series data and improve the accuracy of a data-driven model for combined sewer flow prediction. We considered the sewer system of a small town in Germany as the test case. Precipitation and inflow to the storage tanks are used for the Data-Driven model development. The aim is to predict the flow using precipitation data and examine the impact of data augmentation using synthetic data in model performance. Results show that GAN can successfully generate synthetic time series from real data distribution, which helps more accurate peak flow prediction. However, the model without data augmentation works better for dry weather prediction. Therefore, an ensemble model is suggested to combine the advantages of both models.

## 1    INTRODUCTION

Nowadays, many industry sectors such as health care, automation, financial markets, aerospace, water resources management, and weather forecasts deal with time-series data in their operations and development. Improving the data availability can increase the efficiency of existing infrastructures and foster research for environmental sustainability. Data analysis and mining would facilitate a shift from pure infrastructure development to smart operation and management of our highly interconnected systems with environmental challenges. In recent years, we have witnessed numerous breakthroughs in machine learning techniques in various domains, such as computer vision, language processing, reinforcement learning, and many more [1].

However, still, some fundamental shortcomings hinder progress in environmental management. They include (1) lack of freely available data (e.g., due to data privacy or high expenses of data gathering) for an extended period as well as lack of rare or extreme events in the training data, (2) lack of robust methods for anomaly detection, particularly for drift detection, (3) absence of probabilistic time series forecasting data-driven methods to consider different sources of uncertainty for optimal and robust operation of critical urban water infrastructures, (4) lack of benchmark cases and (5) various sources of uncertainty affecting the data and the resulting models.

Generative algorithms are powerful approaches in data science that can help us overcome the challenges mentioned above in various domains of water and environmental management. GANs are a deep learning architecture for training powerful generator models. The main goal of GANs is to learn from a set of training data and generate new data with the same characteristics as the training data. Initially, they were applied to domains where their results are intuitively assessable, e.g., images. However, GANs have been successfully used to generate time series sequences in the health care, finance, and energy industry and outperform state-of-the-art benchmarks. Nevertheless, GANs' potential in Urban Water Management (UWM) problems are not yet discovered to our knowledge.

In this study, as a proof of concept, we test one primary application of GANs, i.e., data augmentation to urban drainage systems. We use GANs to generate synthetic time series to balance our data set and improve the accuracy of a data-driven combined sewer flow prediction model. As relevant events are relatively rare in the historical data, pure data-driven rainfall-runoff models often underestimate the runoff when predicting these events. The optimal operation of the sewer system during these events, which result in the most critical states for the urban area and environment, depends highly on accurate flow predictions. We used GANs to generate synthetic time series from approximately the same statistical distribution of our data set to overcome these challenges. The generator model enables us to balance our training data with extreme synthetic events. To evaluate the performance of the proposed approach, we train a specific deep neural network model with and without synthetic data and test their performance using a similar test data set. The remainder of this manuscript is structured as follows:

## 2 MATERIALS AND METHODS

### 2.1 Case Study

In this study, the sewer system of a small town in Baden-Württemberg state in Germany is considered to evaluate the performance of the proposed model. Precipitation (mm), temperature (°C), and inflow to the storage tanks are the data set used for the model development. All data are measured from the beginning of June 2017 to the end of January 2018 at a 5-min time resolution [2]. Figure 1 summarizes our data set. It also depicts an example of our measured time series. The aim is to build a black-box simulator to predict combined sewer flow in the system using historical and synthetic data. The model then can be employed for optimal operation, e.g., to minimize the volume and duration of combined sewer overflows (CSOs) of the system using RTC.

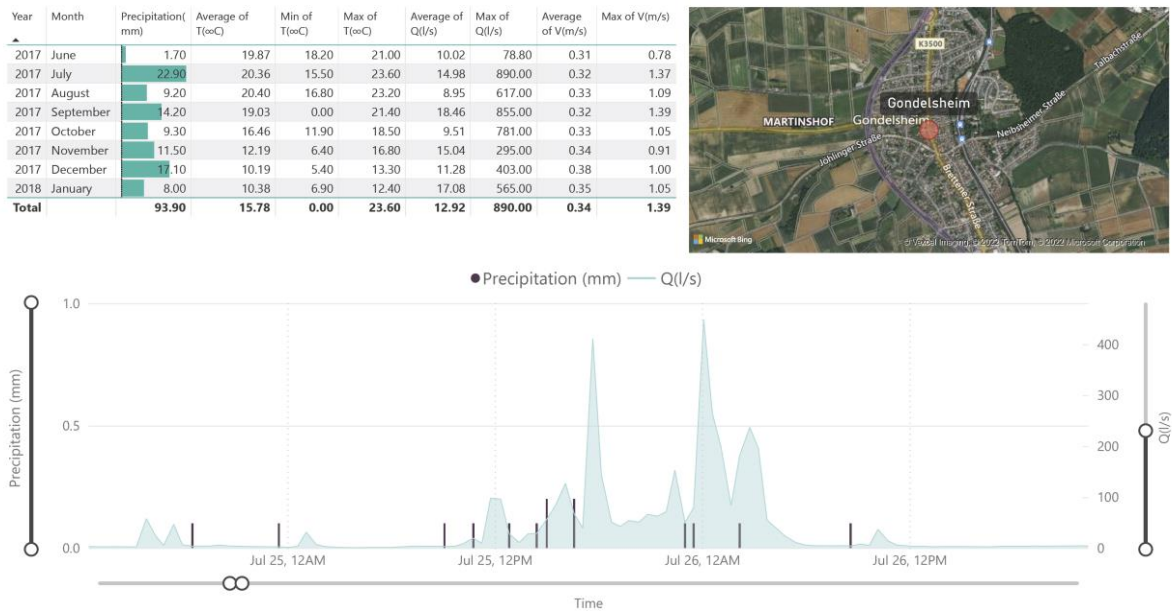

| Year | Month | Precipitation (mm) | Average of T(=C) | Min of T(=C) | Max of T(=C) | Average of Q(l/s) | Max of Q(l/s) | Average of V(m/s) | Max of V(m/s) |
|------|-------|------|------|------|------|------|------|------|------|
| 2017 | June | 1.70 | 19.87 | 18.20 | 21.00 | 10.02 | 78.80 | 0.31 | 0.78 |
| 2017 | July | 22.90 | 20.36 | 15.50 | 23.60 | 14.98 | 890.00 | 0.32 | 1.37 |
| 2017 | August | 9.20 | 20.40 | 16.80 | 23.20 | 8.95 | 617.00 | 0.33 | 1.09 |
| 2017 | September | 4.20 | 19.03 | 0.00 | 21.40 | 18.46 | 855.00 | 0.32 | 1.39 |
| 2017 | October | 9.30 | 16.46 | 11.90 | 18.50 | 9.51 | 781.00 | 0.33 | 1.05 |
| 2017 | November | 11.50 | 12.19 | 6.40 | 16.80 | 15.04 | 295.00 | 0.34 | 0.91 |
| 2017 | December | 27.10 | 10.19 | 5.40 | 13.30 | 11.28 | 403.00 | 0.38 | 1.00 |
| 2018 | January | 8.00 | 10.38 | 6.90 | 12.40 | 17.08 | 565.00 | 0.35 | 1.05 |
| **Total** | | **93.90** | **15.78** | **0.00** | **23.60** | **12.92** | **890.00** | **0.34** | **1.39** |

*Figure 1. Case study*

## 2.2 Problem Description

The release of untreated sewerage into receiving water bodies during rain events can be reduced by dynamically controlling sewer flow and retention volume with sensor networks and automated valves [1]. MPC is an advanced RTC technique in which the optimization is based not only on the knowledge of the system's current state but also on its forecast state. Thus, MPC allows us to improve the monitoring process and optimally utilize the storage capacities of rainwater reservoirs, detention ponds, and in-sewer storage volumes by considering anticipated rainfall, thus, e.g., reducing CSOs [1]. As relevant events are relatively rare in the historical data, pure data-driven rainfall-runoff models often underestimate the runoff when predicting these events. The optimal operation of the sewer system during extreme events, which result in the most critical states for the urban area and environment, depends highly on accurate flow predictions.

## 2.3 Generative Adversarial Networks (GANs)

Generative models are a class of algorithms that aim to generate realistic artificial data. These models approximate the probability distribution of a given data set as closely as possible. Then, we can fabricate new data samples from the approximated distribution. To put it in more concrete terms, assume data set $\mathcal{D}$, which is consists of $N$ i.i.d samples ($x$) from a probability distribution i.e., $\mathcal{D} = \{x_1, \ldots, x_n\}$. We denote these samples as real data and their probability distribution as $P_{real}$. The goal of a generative model is to accurately approximate $P_{real}$ given samples from this distribution i.e., real data. Once we have a good approximation of data distribution, we can sample from this distribution and generate artificial data that follow the real data distribution.

The GAN [3] approach to achieving this goal is to define a mapping function $f$ that transforms samples from a latent space ($P_z$) to the samples in the data space ($P_{real}$). In other words, our goal is to define $f$ as:

$$f(z) = x,$$

where $z \sim P_{latent}$ and $x \sim P_{real}$. Conventionally, GAN utilizes a Gaussian distribution as the latent space. This latent space is also referred to as Noise space in the GAN domain. GAN employs a neural network called Generator ($G$) to approximate $f$. To train $G$, GAN uses another neural network called Discriminator ($D$). The Discriminator is a classifier trained to differentiate between data from the data set (real data) and generated from the generator (fake data). By learning to discriminate between real and fake data, the Discriminator can provide generator feedback regarding how realistic the fake data is. Figure 2 illustrates GAN's structure. First, we sample vector z from latent space $P_z$. Then, generator takes the vector z and maps it to a data sample. Finally, the Discriminator assesses the authenticity of generated with having access to both fake data and generated.

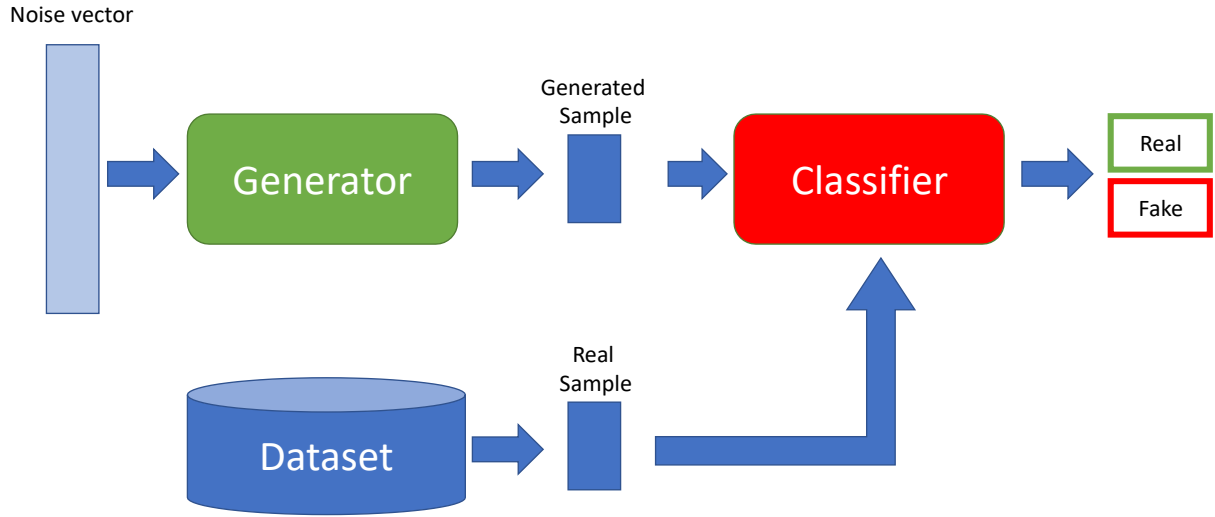

*Figure 2. The abstract representation of GAN architecture*

During the training of the GAN, the Generator and Discriminator are trained interchangeably in an adversarial fashion. We first train Discriminator to classify real and fake data as accurately as possible in one training step. Then, we train the Generator to fabricate the data sample to be identified as real by the Discriminator. In other words, the Generator tries to fool the Discriminator by generating realistic data points while the Discriminator tries not to be fooled by learning the classification between real and fake data. This two-player minimax game is set in motion by optimizing the following value function:

$$\min_{G} \max_{D} V(D, G) = \mathbb{E}_{x \sim P_{real}}[\log(D(x))] + \mathbb{E}_{z \sim P_z}\left[\log\left(1 - D\left(G(z)\right)\right)\right].$$

If both models have enough capacity, the Generator will learn the probability distribution data and $D(x) = \frac{1}{2}$ for all x.

The original GAN proposal is quite challenging to train since the divergences which GANs typically minimize are potentially not continuous with respect to the Generator's parameters [4]. Therefore, various methods have been suggested to improve GAN's training stability. In this work, we adopt the Wasserstein GAN gradient penalty (WGAN-gp) to improve the training process. In WGAN, the training process minimizes the Wasserstein distance between a real distribution and fake distribution, i.e., $W(P_{real}, P_{fake})$ which continues under mild assumptions. Using Kantorovich-Rubinstein duality [5], the value function of WGAN is defined as:

$$\min_{G} \max_{D \in \mathcal{L}} = \mathbb{E}_{x \sim P_{real}}[D(x)] - \mathbb{E}_{\overline{x} \sim P_{fake}}[D(\overline{x})],$$

where, $\mathcal{L}$ is the set of 1-Lipschitz functions and $P_{fake}$ is the distribution learned by the Generator implicitly. In this case, the discriminator network is replaced with the critic network. The critic aims to assess the authenticity of its input and assigns a numerical score based on the similarity of the input to real data. The WGAN-gp [6] imposes a penalty on gradient norm to enforce the Lipschitz constraint. Hence, the final objective function is:

$$L = E_{\overline{x} \sim P_{fake}}[D(\overline{x})] - E_{x \sim P_{real}}[D(x)] + \lambda \, \mathbb{E}_{\hat{x} \sim P_{\hat{x}}}\left[\left(|\nabla_{\hat{x}} D(\hat{x})|_{\{2\}} - 1\right)^2\right]$$

where $\lambda$ is the penalty coefficient and $P_{\hat{x}}$ is implicitly defined by sampling uniformly on lines between patis of points sampled from $P_{real}$ and $P_{fake}$.

## 2.4 GAN Architecture

Figure 3 and 4 illustrate the Generator and Discriminator structures, respectively. The Generator receives the start token and the noise vector from latent space. The start token is the mean value of each data channel and serves as a common starting point for our generation. Note that the start token is not part of the final generation. The Generator uses the start token as input of a GRU layer and employs the noise vector as an initial hidden state of the GRU layer. Then, the output of GRU passes through a fully connected layer to produce the first step of our fabricated time series. Next, we feed the generated time-step back to the generator to fabricate the next time step. This auto-regressive process continues until we generate the desired number of time-step (24 time steps in this scenario). The Discriminator receives a time frame from the generator or dataset and passes it through a GRU layer. Then, the output of the GRU layer is fed to a fully connected layer to obtain the final output of Discriminator.

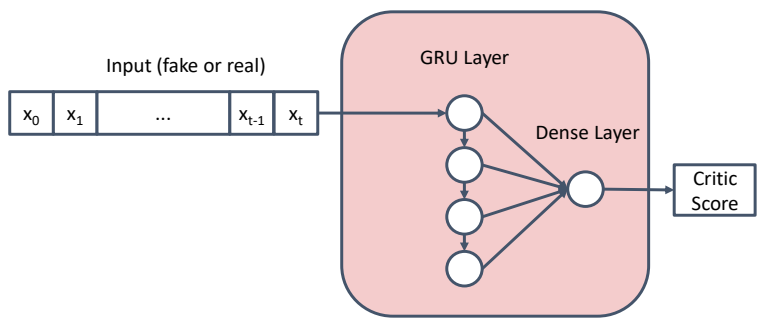

*Figure 3. The architecture of critic*

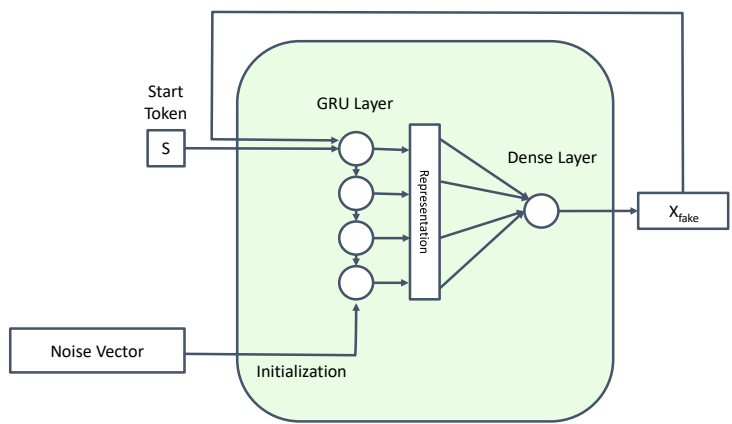

*Figure 4. The architecture of generator*

# 3    RESULTS AND DISCUSSION

## 3.1    Pre-processing pipeline

In this paper, we aim to generate artificial data to enhance the size of the dataset at hand, especially for rare situations where data sample scarcity makes it difficult for any model to learn the data pattern. Figure 5 illustrates our data preprocessing pipeline. Since the flow channel had a skewed distribution toward zero value (Figure ), first, we applied log transform on this channel to obtain a less skewed distribution. Then, we standardized data distribution by a linear transformation to have mean=0 and standard deviation = 1. The next step of preprocessing the data turned into a series of time-frames using the rolling window technique with windows size = 24.

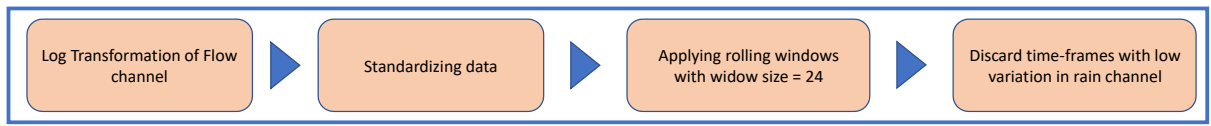

*Figure 5: The preprocessing pipeline*

The rain channel data consists of long periods of dry days (low variation section) and a few short periods of rainy days (high variations). Our goal is to generate more rare cases (rainy days) to augment our data set. Therefore, we discard those data frames which did not have any variation in their time frame.

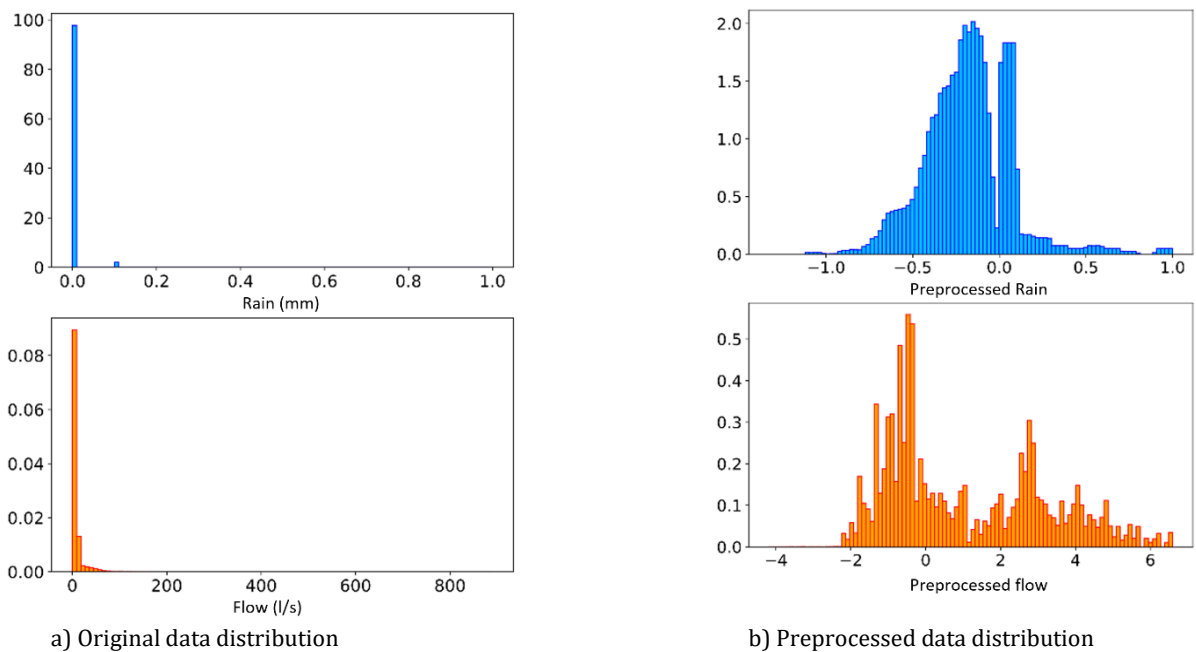

a) Original data distribution                    b) Preprocessed data distribution

*Figure 6: The data distribution before and after preprocessing*

## 3.2 Experiment Set-up

In order to train a GAN successfully, we need to select the Discriminator and the Generator hyperparameters carefully. If one of the networks improves substantially faster than the other network during training, it is likely that training stops before reaching equilibrium. Hence, we perform a hyperparameter tuning for the Generator and the Discriminator. Table x presents the list of hyperparameters we tuned alongside the search space for each hyperparameter and the final value we select for the corresponding hyperparameter.

During training, the value function of GAN informs us about the performance of the Generator against the current Discriminator for a given batch. It does not provide any information regarding the performance of the Generator in general. Hence, we need a secondary assessment method to determine the best-performing model during the hyperparameter tuning process. The task of assessing a generative model in the time-series domain is a problem, and there is not any suitable measurement method for this task currently. For this work, we employed Jensen-Shannon Divergence (JSD) between the batch of real points from the dataset and the batch of fake points from the Generator.

*Table 1: This table presents the list of GAN's hyperparameters alongside the search of space of each hyperparameter and the best hyperparameter we found during hyperparameter tuning*

| Hyperparameter | Search space | Best hyperparameter |
|----------------|--------------|---------------------|
| GRU Layers in G | [1,4] | 3 |
| GRU layers in D | [1,4] | 3 |
| GRU hidden size in G | [32, 512] | 450 |
| GRU hidden size in D | [32, 512] | 120 |

$$JSD(P_{real}||P_{fake}) = \frac{1}{2}KLD(P_{real}||M) + \frac{1}{2}KLD(P_{fake}||M),$$

where $M = \frac{1}{2}(P_{real} + P_{fake})$ and KLD is Kullback–Leibler divergence:

$$KLD(P||Q) = \sum_{x \in \chi} P(x) \log\left(\frac{P(x)}{Q(x)}\right),$$

where $\chi$ is the probability space. JSD does not consider the auto-correlation between the consecutive point in time series or the channels' correlation. However, it provides us with an easy to compute assessment method that can weakly measure the performance of GAN during the training and allows us to prune poor-performing configurations in the early stages. Then, we can select the best model based on the improvement of downstream tasks (in this case, forecasting) from the set of good-performing candidates.

For hyperparameter tuning, we utilized Bayesian Optimization Hyper Band Algorithm (BOHB) [7]. For implementing the networks, we used Pytorch [8] and for performing hyperparameter tuning, we employed RayTune package [9]. All experiments were executed on two Nvidia RTX 1080Ti graphic cards.

### 3.3    Data Augmentation with GAN

The best model that is obtained during hyperparameter tuning has JSD equal to 0.33. Figure 7 shows the generated time series.

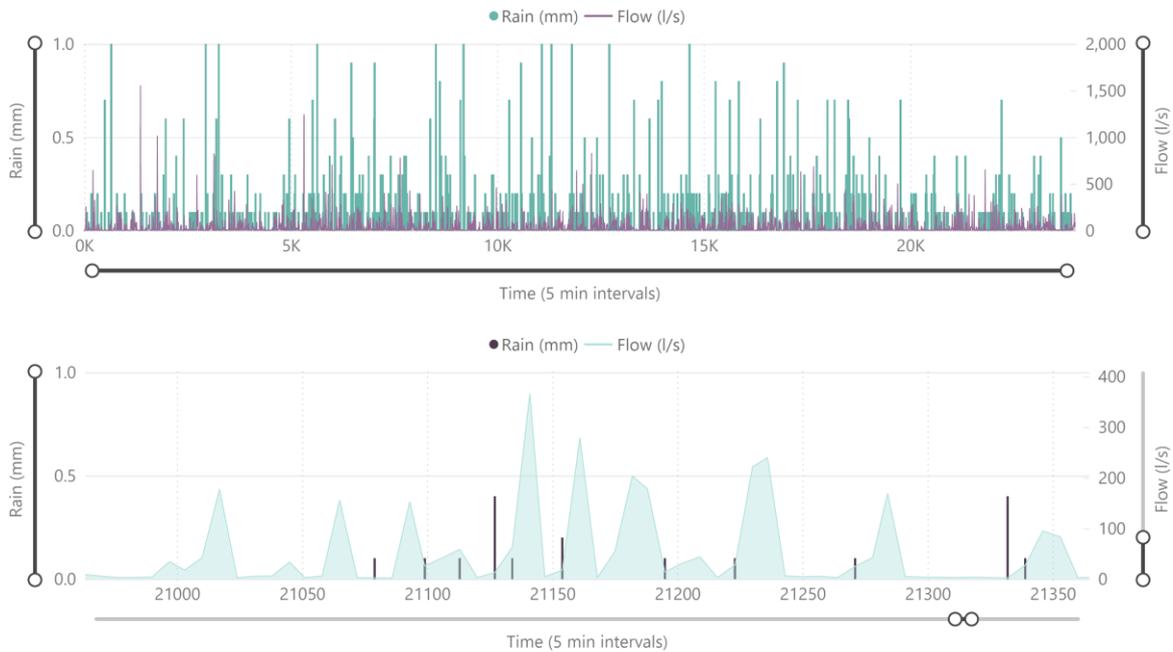

*Figure 7: Synthetic time series represented in different resolutions*

### 3.4    Data-Driven Forecasting Model

For training the data-driven forecasting model, we transformed the time series prediction problem into a supervised learning problem. We split the rainfall time series into 24-time steps (120 minutes) windows. The aim is to predict flow using these windows, as depicted in figure 8.

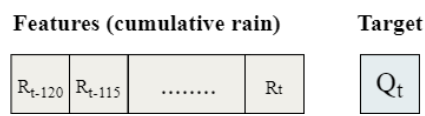

*Figure 8: Splitting rainfall time series into 24-time steps windows for supervised learning*

We trained two different models to solve this problem, one with synthetic data and one without. 8000 windows generated from the synthetic data set are used to balance the training data for the first model. We also balance the second model using the oversampling method for better comparison. The train data set for both models consist of 38000 windows from the real dataset. The test data set has 13000 windows only from the real data set. We built two sequential models in TensorFlow with some LSTM layers connecting to some Dense layers. ADAM method is used as an optimizer and means average error (MAE) as the cost function. A simple grid search is used to tune the hyperparameters, including the number of LSTM and Dense layers and their neurons and batch size. The results show that the model without synthetic data performs slightly better than the model with synthetic data.  The MAE for the model without synthetic data is 5.01, while this value is 5.40 for the model with synthetic data. However, a closer look at the results reveals that the model with synthetic data has higher accuracy in predicting peak values which are the main concerns of this study. Figure 9 presents the results of both models for the test dataset. As it can be seen, the model with synthetic data predicts the peak values with higher accuracy for all extreme events. Still, the other model has better accuracy for dry weather conditions.

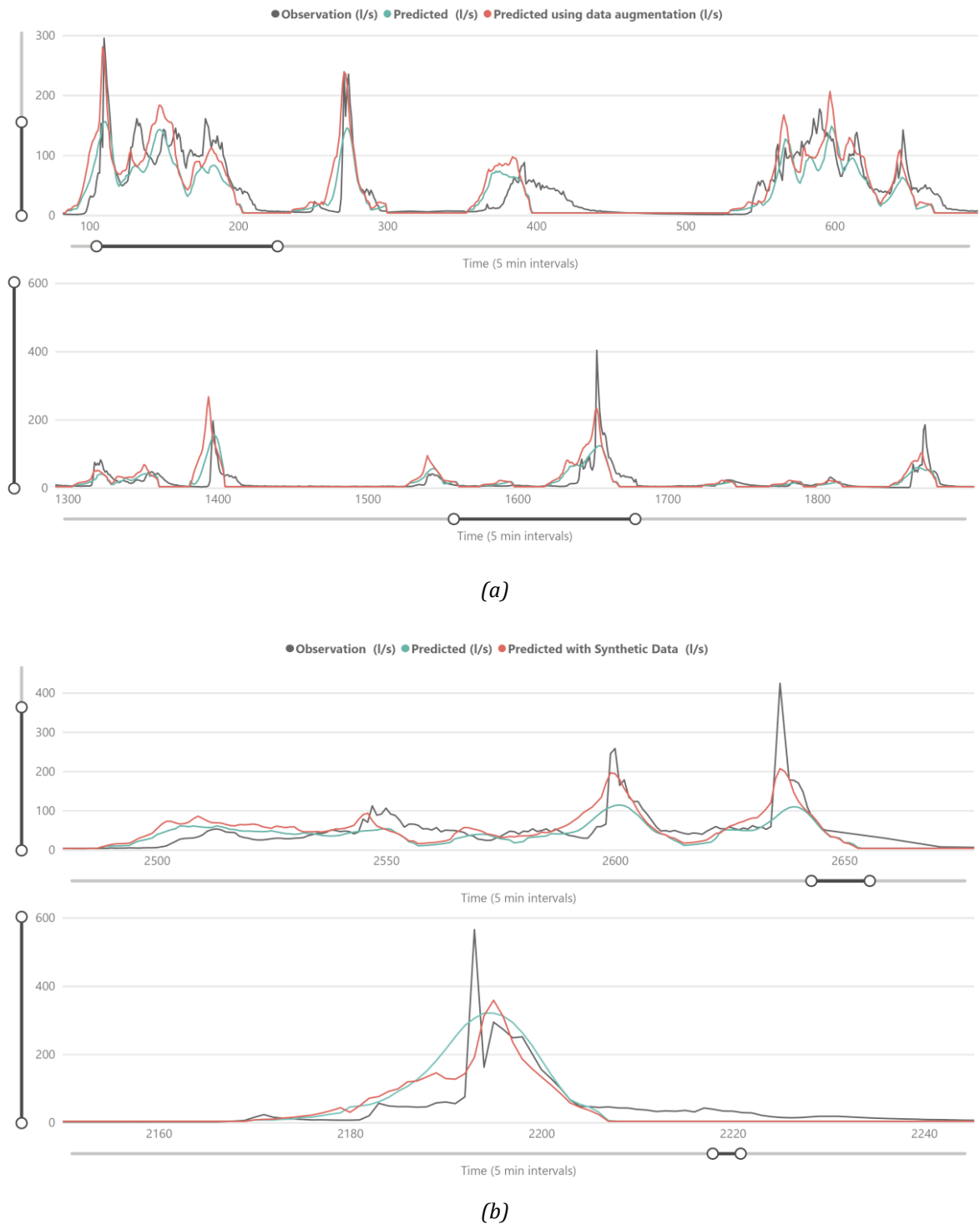

*(a)*

*(b)*

*Figure 9: comparing flow predictions with observations for models with and without synthetic data*

## 4    CONCLUSION

As moderate or extreme rain events are relatively rare in the historical data, pure data-driven rainfall-runoff models often underestimate the runoff when predicting these events. The optimal operation of the sewer system during extreme events, which result in the most critical states for the urban area and environment, depends highly on accurate flow predictions. To overcome these challenges, we used Generative Adversarial Networks (GANs). GANs are a class of methods capable of learning data distribution to generate artificial data. We developed a GAN model to generate synthetic time series to balance our limited and imbalanced recorded time series data to improve the accuracy of a data-driven model for combined sewer flow prediction using historical rainfall and measured flow data.

Results show that balancing the training dataset using the synthetic data generated by the developed GAN improves the accuracy of the data-driven model in peak flow prediction. However, this data augmentation method reduces the model performance in dry weather conditions. Some potential research themes to complement, expand, and build upon the presented study are given in the following.

- Clustering time series into dry and wet weather series and using different data-driven models for each series

- Using more complicated models like physic-informed ML

- Ensemble learning

- In the future, GANs can be applied for probabilistic anomaly detection and missing data imputation in urban water management [10]

- GANs can be used  also for Probabilistic rainfall-runoff modeling for model predictive combines sewer overflow control [11]